\documentclass[sigconf,screen]{acmart}

\usepackage[utf8]{inputenc} 
\usepackage[T1]{fontenc}    
\usepackage{hyperref}       
\usepackage{url}            
\usepackage{booktabs}       
\usepackage{amsfonts}       
\usepackage{nicefrac}       
\usepackage{microtype}      
\usepackage{graphicx}
\usepackage{enumitem}
\usepackage{tabu}
\usepackage{mathtools}
\usepackage{amsmath}
\usepackage{array}
\usepackage{amsmath,amsfonts,amssymb}
\usepackage{graphicx}
\usepackage{fancyhdr}
\usepackage{makecell}
\usepackage{multirow}
\usepackage{microtype}
\usepackage{balance}
\usepackage{natbib}
\usepackage{graphicx}
\usepackage{amsmath, amsthm, amssymb}
\usepackage{array}
\usepackage{tikz}
\usepackage{bussproofs}
\usepackage[utf8]{inputenc}

\usepackage{caption} 
\usepackage{booktabs}

\usepackage{algorithm}
\usepackage{algpseudocode}

\newcommand{\firstl}{Gated Multimodal Embedding\ }
\newcommand{\firsts}{GME}
\newcommand{\lastl}{LSTM with Temporal Attention\ }
\newcommand{\lasts}{LSTM(A)}
\newcommand{\noatt}{LSTM}
\newcommand{\ourl}{\firstl \lastl\ }
\newcommand{\ours}{\firsts-\lasts}

\makeatletter
\def\@fnsymbol#1{\ensuremath{\ifcase#1\or *\or *\or *\or
   \mathsection\or \mathparagraph\or \|\or **\or \dagger\dagger
   \or \ddagger\ddagger \else\@ctrerr\fi}}
\makeatother

\begin{document}
\copyrightyear{2017}
\acmYear{2017} 
\setcopyright{acmlicensed}
\acmConference[ICMI'17]{19th ACM International Conference on Multimodal Interaction}{November 13--17, 2017}{Glasgow, UK}\acmPrice{15.00}\acmDOI{10.1145/3136755.3136801}
\acmISBN{978-1-4503-5543-8/17/11}

\title{Multimodal Sentiment Analysis with Word-Level Fusion and Reinforcement Learning}

\author{Minghai Chen}
\authornote{Equal contribution.}
\affiliation{%
  \institution{Language Technologies Institute \\ Carnegie Mellon University}
  \city{Pittsburgh} 
  \state{PA}
  \postcode{15213}
  \country{USA}
}
\email{minghai1@cs.cmu.edu}

\author{Sen Wang}
\authornotemark[1]
\affiliation{%
  \institution{Language Technologies Institute \\ Carnegie Mellon University}
  \city{Pittsburgh}
  \state{PA}
  \country{USA}
  \postcode{15213}
}
\email{senw1@cs.cmu.edu}

\author{Paul Pu Liang}
\authornotemark[1]
\affiliation{%
  \institution{Language Technologies Institute \\ Carnegie Mellon University}
  \city{Pittsburgh} 
  \state{PA} 
  \postcode{15213}
  \country{USA}
}
\email{pliang@cs.cmu.edu}

\author{Tadas Baltru\v{s}aitis}
\affiliation{%
  \institution{Language Technologies Institute \\ Carnegie Mellon University}
  \city{Pittsburgh} 
  \state{PA} 
  \postcode{15213}
  \country{USA}
}
\email{tb346@cl.cam.ac.uk}
\author{Amir Zadeh}
\affiliation{%
  \institution{Language Technologies Institute \\ Carnegie Mellon University}
  \city{Pittsburgh} 
  \state{PA} 
  \postcode{15213}
  \country{USA}
}
\email{abagherz@cs.cmu.edu}
\author{Louis-Philippe Morency}
\affiliation{%
  \institution{Language Technologies Institute \\ Carnegie Mellon University}
  \city{Pittsburgh} 
  \state{PA} 
  \postcode{15213}
  \country{USA}
}
\email{morency@cs.cmu.edu}

\renewcommand{\shortauthors}{Chen, Wang, Liang, Baltru\v{s}aitis, Zadeh and Morency}
\title[{Multimodal Sentiment Analysis, Word-Level Fusion, Reinforcement Learning}]{Multimodal Sentiment Analysis with Word-Level Fusion and Reinforcement Learning}

\begin{abstract}
With the increasing popularity of video sharing websites such as YouTube and Facebook, multimodal sentiment analysis has received increasing attention from the scientific community. Contrary to previous works in multimodal sentiment analysis which focus on holistic information in speech segments such as bag of words representations and average facial expression intensity, we develop a novel deep architecture for multimodal sentiment analysis that performs modality fusion at the word level. In this paper, we propose the \ourl(\ours) model that is composed of 2 modules. The \firstl alleviates the difficulties of fusion when there are noisy modalities. The \lastl performs word level fusion at a finer fusion resolution between input modalities and attends to the most important time steps. As a result, the {\ours } is able to better model the multimodal structure of speech through time and perform better sentiment comprehension. We demonstrate the effectiveness of this approach on the publicly-available Multimodal Corpus of Sentiment Intensity and Subjectivity Analysis (CMU-MOSI) dataset by achieving state-of-the-art sentiment classification and regression results. Qualitative analysis on our model emphasizes the importance of the Temporal Attention Layer in sentiment prediction because the additional acoustic and visual modalities are noisy. We also demonstrate the effectiveness of the \firstl in selectively filtering these noisy modalities out. Our results and analysis open new areas in the study of sentiment analysis in human communication and provide new models for multimodal fusion. 
\end{abstract}


\maketitle

\section{Introduction}
Multimodal sentiment analysis is an emerging field at the intersection of natural language processing, computer vision, and speech processing. Sentiment analysis aims to find the attitude of a speaker or writer towards a document, topic or an event \citep{pang2008opinion}. Sentiment can be expressed by the spoken words, the emotional tone of the delivery and the accompanying facial expressions. As a result, it is helpful to combine visual, language, and acoustic modalities for sentiment prediction \citep{perez-rosas_utterance-level_2013}. To combine cues from different modalities, previous work mainly focused on holistic video-level feature fusion. This was done with simple features (such as bag-of-words and average smile intensity) calculated over an entire video as representations of verbal, visual and acoustic features \citep{wang2016select,zadeh2016multimodal}. These simplistic fusion approaches mostly ignore the structure of speech by focusing on simple statistics from videos and combining modalities at an abstract level. 

The cornerstone of our approach is capturing the full structure of speech using a time-dependent recurrent approach that is able to properly perform modality fusion at every timestep. This understanding of speech structure is important due to two major reasons: 1) There are \textit{local} interactions between modalities, such as how loud a word is being uttered which has roots in language and acoustic modalities or whether or not a word was accompanied by a smile which has roots in language and vision. Considering local interaction helps in dealing with commonly researched problems in natural language processing such as ambiguity, sarcasm or limited context by providing more information from the visual and acoustic modalities. Consider the word ``crazy''; this word can have a positive sentiment if accompanied by a smile or can have a negative sentiment if accompanied by a frown. At the same time the word ``great'' accompanied by a frown implies sarcastic speech. Also, in limited context, inference of sentiment intensity is difficult. For example the word ``good'' accompanied by neutral nonverbal behavior could mean that the utterance is positive; but the same word accompanied with big smile could mean highly positive sentiment. 2) There are \textit{global} interactions between modalities mostly established based on temporal relations between modalities. Examples include a delayed laughter due to a speaker's utterance of words or a delayed smile because of a speech pause. Each of the modalities also have their own intramodal interactions (such as how different gestures happen over the utterance), which can be characterized as the global structure of speech.

To properly model structure of speech, two key questions need to be answered: ``what modality to look for at each moment in time?'' and ``what moments in speech are important in the communication?''. To address the first question, a model should be able to ``gate'' the useful modalities at each moment in time. If a modality does not contain useful information or the modality is too noisy that negatively affects the performance of the model, the model should be able to shut off the modality and perform inference based on information present in the other modalities. To address the second question, a model should be able to divert it's ``attention'' to key moments of communication such as when a polarized word is uttered or when a smile happens. In this paper we introduce the Gated Multimodal Embedding LSTM with Temporal Attention model, which explicitly accounts for these two key questions by using a gated mechanism for multimodal fusion at each time step and a Temporal Attention Layer for sentiment prediction. Our model is able to explore the structure of speech through a stateful recurrent mechanism and perform fusion at word level between different modalities. This gives our model the ability to account for local (by word level fusion of multimodal information) and global (by stateful multimodal memory) interactions between modalities. 

The remainder of the paper is as follows. In Section \ref{Related Work}, we review the related work in multimodal sentiment analysis. In Section \ref{Proposed Approaches}, we give formal definition of our problems and present our approaches in detail. In Section \ref{Experimental Methodology}, we describe the CMU-MOSI dataset, experimental methodology and baseline models. The results on CMU-MOSI dataset are presented in Section \ref{Results}. A detailed analysis of our model's components is provided in Section \ref{Discussion}. Section \ref{Conclusion} concludes the paper.
\section{Related Work}
\label{Related Work}
Deep learning approaches have been became extremely popular in the past few years \cite{young2017recent}. The field of multimodal machine learning specifically has gotten unprecedented momentum \cite{baltruvsaitis2017multimodal}. Multimodal models have been used for sentiment analysis \cite{zadeh2016multimodal,poria2017ensemble,zadeh2016mosi}; medical purposes, such as detection of suicidal risk, PTSD and depression \cite{scherer_self-reported_2016, venek2016adolescent, yu2013multimodal, valstar2016avec}; emotion recognition \cite{poria2017review}; image captioning and media description \cite{you2016image, donahue2015long}; question answering \cite{antol2015vqa}; and multimodal translation \cite{specia2016shared}. Our work is specifically connected to the following areas: 

\textbf{Sentiment analysis} on written text modality has been well-studied \citep{pang2008opinion} with models that predict sentiment from language. Early works used the bag of words or n-gram representations \citep{yang2012extracting} to derive sentiment from individual words. Other approaches focused on opinionated words \citep{taboada2011lexicon, poria2012fuzzy,poria2016sentic}, and some applied more complicated structures such as trees \citep{socher2013recursive} and graphs \citep{poria2014dependency}. These structures aimed to derive the sentiment of sentence based on the sentiment of individual words and their compositions. More recent works used dependency-based semantic analysis \citep{agarwal2015concept}, distributional representations for sentiment \citep{iyyer2015deep} and a convolutional architecture for the semantic modeling of sentences \citep{kalchbrenner2014convolutional}. However, we are primarily working on spoken text rather than written text, which gives us the opportunity to integrate additional audio and visual modalities. These modalities are helpful by providing additional information, but sometimes may be noisy. 

\textbf{Multimodal sentiment analysis} integrates verbal and nonverbal behavior to predict user sentiment. Though various multimodal datasets with sentiment labels exist \citep{wollmer2013youtube, morency2011towards, perez2013utterance}, the CMU-MOSI dataset \citep{zadeh2016multimodal} is the first dataset with opinion level sentiment labels.

Recent multimodal sentiment analysis approaches focus on deep neural networks, including Convolutional Neural Networks (CNN) with multiple-kernel learning \citep{poria2015deep} and Select-Additive Learning (SAL) CNN \citep{wang2016select} which learns generalizable features across speakers. \citep{behnaz2016dnn} uses an unimodal deep neural network for three modalities (language, acoustic and visual) and explores the effectiveness of early fusion and late fusion. The features of the three modalities are similar to our work, but to fuse the multimodal information their work simply uses an concatenation approach at video level while we propose and justify the use of more advanced methods for multimodal fusion.

Besides sentiment, other speakers' attributes such as persuasion, passion and confidence could also be analyzed by similar methods \citep{Park:2014:CAP:2663204.2663260,majumder2017deep}. \citep{Chatterjee_etAl2015} proposes an ensemble classification approach that combines two different perspectives on classifying multimodal data: the first perspective assumes inter-modality conditional independence, while the second perspective explicitly captures the correlation between modalities is  and recognized by a clustering based kernel similarity approach. These methods can also be applied to multimodal sentiment analysis.

Our gated controller for visual and acoustic modality is inspired by the controller used by Zoph and Le \citep{zoph2016neural}, where they use a Recurrent Neural Network (RNN) controller to determine the structure of a CNN.

Compared to previous work, our method has two major advantages. To the best of our knowledge, our model is the first to use word level modality fusion, which means that we align each word to corresponding video frames and audio segments. Secondly, we are also the first to propose an attention layer and a input gate controller trained by reinforcement learning to approach the problem of noisy modalities.
\section{Proposed Approaches}
\label{Proposed Approaches}
In this section, we give a detailed description of our proposed approach which will be divided into 2 modules: the \firstl Layer and the \lastl model. The \firstl Layer performs selective multimodal fusion at each time step (word level) using input modality gates, and the \lastl performs sentiment prediction with attention to each time step. Together, these modules combine to form the \ourl (\ours). The section ends with the training details for the \ours.

\begin{figure*}
\begin{center}
\includegraphics[width=0.7\textwidth]{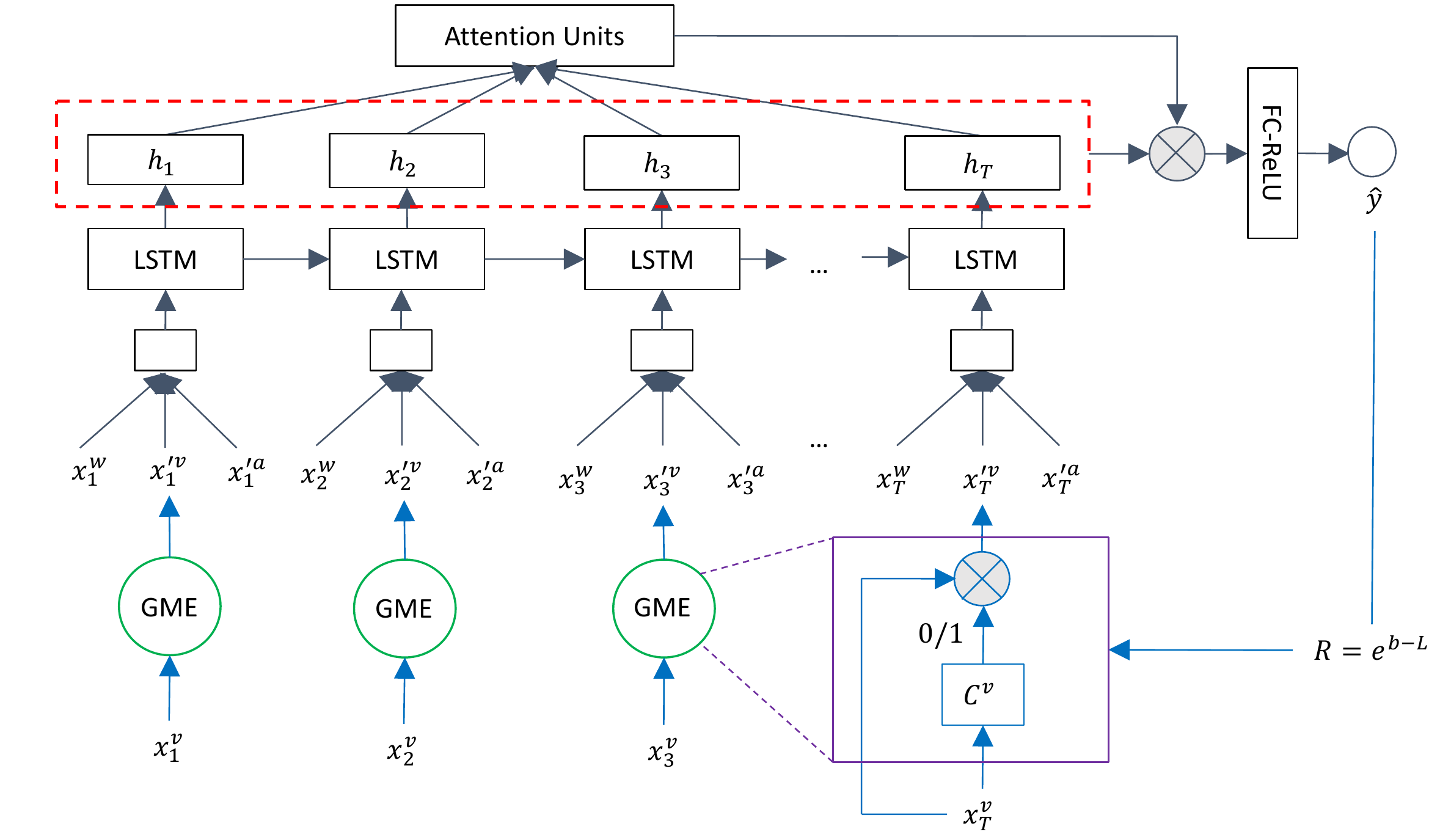}
\caption{Architecture of the {\ours } model for the visual modality. $C^v$ is the controller for the visual modality that selectively allows visual inputs $\mathbf{x}^v_t$ to pass. FC-ReLU is a fully-connected layer with rectified linear unit (ReLU) as activation. After obtaining a sentiment prediction $\hat{y}$ and loss $\mathcal{L}$, we use $R = e^{b-\mathcal{L}}$ as the reward signal to train the visual input gate controller $C^v$.}
\label{gate}
\end{center}
\end{figure*}

\subsection{Gated Multimodal Embedding}

The first component is the Gated Multimodal Embedding that performs multimodal fusion by learning the local interactions between modalities. Suppose the dataset contains $N$ video clips, each containing an opinion mapped to sentiment intensity. A video clip contains $T$ time steps, where each time step corresponds to a word. Each video clip is also labeled with the ground truth sentiment value $y \in \mathbb{R}$. We align words with their corresponding video and audio segment using the Penn Phonetics Lab Forced Aligner (P2FA) \citep{P2FA}. P2FA is a software that computes an alignment between a speech audio file and a verbatim text transcript. Following the alignment, at each word level time step $t$, we obtain the aligned feature vectors for language (word), acoustic and visual modalities: $\mathbf{x}_{t}^w$, $\mathbf{x}_{t}^a$, $\mathbf{x}_{t}^v$ respectively. 

One problem of previous models is that of \textbf{noisy modalities} during multimodal fusion since the textual modality may be negatively affected by the visual and audio modalities. As a result, useful textual features may be ignored because the corresponding visual or acoustic feature is noisy and important information may be lost. 

Our solution is to introduce an on/off input gate controller that determines if the acoustic or visual modality at each time step should contribute to the overall prediction. The reason why we apply the input gate controller on acoustic/visual features while always letting textual features in is that the language modality is much more reliable for multimodal sentiment analysis than visual or acoustic. Also visual and acoustic modalities can be noisy or unreliable since audio visual feature extraction is done automatically using methods that add additional noise. 

Mathematically, we formalize this with inputs $\mathbf{x}^a_t$ and $\mathbf{x}^v_t$ representing the audio and visual inputs at time step $t$ respectively. We have a controller $C^a$, with weights $\theta_a$, for determining the on/off of audio modalities, and $C^v$, with weights $\theta_v$, for determining the on/off of visual modalities. These controllers are implemented as deep neural networks $C^a (\ \cdot \ ;\theta_a)$ and $C^v (\ \cdot \ ;\theta_v)$ that take in $\mathbf{x}^a_t$ and $\mathbf{x}^v_t$ as input and outputs a binary label $c^a_t$ and $c^v_t$ (0/1) respectively. The binary output of these controllers mimics the act of accepting or rejecting a modality based on its noise level. The new inputs $\mathbf{x}'^a_t$ and $\mathbf{x}'^v_t$ to the network are:
\begin{eqnarray}
\mathbf{x}'^a_t = c^a_t \cdot \mathbf{x}^a_t =  C^a(\mathbf{x}^a_t ; \theta_a) \cdot \mathbf{x}^a_t\\
\mathbf{x}'^v_t = c^v_t \cdot \mathbf{x}^v_t = C^v(\mathbf{x}^v_t ; \theta_a) \cdot \mathbf{x}^v_t
\end{eqnarray}

We concatenate features from three different modalities: visual, audio and text to form the inputs $\mathbf{x}_t$ to the word level LSTM with Temporal Attention, described in the next section. By extracting features with alignment at the word level, we exploit the temporal correlation among different modalities
$\mathbf{x}_t^w, \mathbf{x}'^a_t, \mathbf{x}'^v_t$, but at the same time the model is less affected by the impact of noisy modalities during multimodal fusion.
\begin{eqnarray}
\mathbf{x}_t &=& \begin{bmatrix}
         \mathbf{x}_t^w \\
         \mathbf{x}'^a_t \\
         \mathbf{x}'^v_t
        \end{bmatrix}
\end{eqnarray}

\subsection{\lastl}

The second component is a sentiment prediction model that captures the temporal interactions on the multimodal embedding layer. This component learns the global interactions between modalities for sentiment prediction. To do so, we use an \lastl (\textbf{\lasts}). The \firstl $\mathbf{x}_t$ is passed as input to the LSTM at each time step $t$. A LSTM \citep{Hochreiter:1997:LSM:1246443.1246450} with a forget gate \citep{gers2000learning} is used to learn global temporal information on multimodal input data $\mathbf{X}_t$:
\begin{eqnarray}
\begin{pmatrix}
         \mathbf{i}\\
         \mathbf{f}\\
         \mathbf{o} \\
         \mathbf{m}
        \end{pmatrix} &=& \begin{pmatrix}
         sigmoid\\
         sigmoid\\
         sigmoid \\
         tanh 
        \end{pmatrix} \mathbf{U} \begin{pmatrix}
         \mathbf{X}_t \mathbf{W}\\
         \mathbf{h}_{t-1}
        \end{pmatrix}\\
\mathbf{c}_{t} &=& \mathbf{f} \odot \mathbf{c}_{t-1} + \mathbf{i} \odot \mathbf{m} \\
\mathbf{h}_t &=& \mathbf{o} \odot tanh(\mathbf{c}_t)
\end{eqnarray}
where $\mathbf{i}$, $\mathbf{f}$ and $\mathbf{o}$ are the input, forget and output gates of the LSTM, $\mathbf{c}$ is the LSTM memory cell and $\mathbf{h}$ is the LSTM output, $\mathbf{W}$ and $\mathbf{U}$ linearly transform $\mathbf{X}_t$ and $\mathbf{h}_{t-1}$ respectively into the LSTM gate space and the $\odot$ operator denotes the Hadamard product (entry-wise product).

We use an attention model similar to \citep{wangattention} to selectively combine temporal information from the input modalities by adaptively predicting the most important time steps towards sentiment of a video clip. We expect relevant information to sentiment to have high attention weights. For example if a person is ``crying'' or ``laughing'', this information is relevant to the sentiment of the opinion and should have higher importance than a neutral word such as ``movie''. This attention mechanism also allows the modalities to act as complimentary information. In cases where language is not helpful, the model can adaptively focus on the presence of sentiment related non-verbal behaviors such as facial gestures and tone of voice.

Mathematically, we add a soft attention layer $\mathbf{\alpha}$ on top of the sequence of LSTM hidden outputs. $\mathbf{\alpha}$ is obtained by multiplying the hidden layer at each time step $t$ with a shared vector $\mathbf{w}$ and passing the sequence through a softmax function. 
\begin{eqnarray}
\mathbf{\alpha}&=& softmax \begin{pmatrix} \begin{bmatrix}
         \mathbf{w}^\top \mathbf{h}_1 \\
         \cdot \\
         \cdot \\
         \cdot \\
         \mathbf{w}^\top \mathbf{h}_T
        \end{bmatrix} \end{pmatrix}
\end{eqnarray}
The attention units $\alpha$ are used to weight the importance of each time step's hidden layer to final sentiment prediction. Suppose $\mathbf{H}$ represents the matrix of all hidden units of the LSTM $[\mathbf{h}_1; ...; \mathbf{h}_T]$. Then the final sentiment prediction $\hat{y}$ is obtained by:
\begin{eqnarray}
\mathbf{z}&=&\mathbf{H}\mathbf{\alpha} \\
\hat{y} &=&  Q(z)
\end{eqnarray}
where function $Q$ represents a dense layer with a non-linear activation.
We select Mean Absolute Error (MAE) as the loss function. Though Mean Square Error (MSE) is a more popular choice for loss function, MAE is a common criteria for sentiment analysis \citep{zadeh2016mosi}.
\begin{eqnarray}
\mathcal{L} &=& \frac{1}{N} \sum_{i=1}^N |\hat{y}_i - y_i|
\end{eqnarray}
Figure \ref{gate} shows the full structure of the {\ours } model.

\subsection{Training Details for \ours}

To train the \ours, we need to know how output decisions of the controller affects the performance of our {\lasts } model. Given the weights of the gate controller and input data $\mathbf{x}^a_{1:T}$ and $\mathbf{x}^v_{1:T}$, the controller decides whether we should reject an input or not. The rejected inputs are replaced with $\mathbf{0}$, while the accepted inputs are not changed. In this way we obtain the new inputs $\mathbf{x}'^a_{1:T}$ and $\mathbf{x}'^v_{1:T}$. After we train the {\lasts } with the new inputs $(\mathbf{x}^w_{1:T}, \mathbf{x}'^a_{1:T}, \mathbf{x}'^v_{1:T})$, we get a MAE loss, $\mathcal{L}$, on the validation set. Here $\mathcal{L}$ can be seen an indicator of how well our controller affects the performance of the model. Note that that lower MAE implies better performance, so we use $e^{-\mathcal{L}}$ as the reward signal to train the controllers.

Take the visual controller $C^v$ as an example: we are maximizing the expected reward, represented by $J(\theta^v)$:
\begin{eqnarray}
	J(\theta^v) = \mathbf{E}_{P(c^v_{1:T}|	\mathbf{x}^v_{1:T}; \theta^v)}[e^{-\mathcal{L}}]
\end{eqnarray}
where $T$ is the total number of time steps in the dataset. The sentiment prediction MAE $\mathcal{L}$ in the reward signal is non-convex and non-differentiable with respect to the parameters of the {\firsts } since changes in the outputs of the GME change the MAE $\mathcal{L}$ in a discrete manner. Straightforward gradient descent methods will not explore all the possible regions of the function. This form of problem has been studied in reinforcement learning where policy gradient methods balance exploration and optimization by randomly sampling many possible outputs of the {\firsts } controller before optimizing for best performance. Specifically, the REINFORCE algorithm \citep{Williams1992} is used to iteratively update $\theta^v$:
\begin{eqnarray}
	\nabla_{\theta^v} J(\theta^v) = \sum_{i=1}^{T}\mathbf{E}_{P(c^v_{1:T}| \mathbf{x}^v_{1:T}; \theta^v)}[\nabla_{\theta^v}\log P(c^v_i|\mathbf{x}^v_{i}; \theta^v) e^{-\mathcal{L}}]
\end{eqnarray}
An empirical approximation of the above quantity is to sample the outputs of the controller \citep{zoph2016neural}:
\begin{eqnarray}
	\nabla_{\theta^v} J(\theta^v) \approx \frac{1}{n} \sum_{k=1}^{n}\sum_{i=1}^{T}\nabla_{\theta^v}\log P(c^v_i|\mathbf{x}^v_{i}; \theta^v) e^{-\mathcal{L}_k}
\end{eqnarray}
where $n$ is the number of different inputs datasets $(\mathbf{x}^w_{1:T}, \mathbf{x}'^a_{1:T}, \mathbf{x}'^v_{1:T})$ that the controller samples, and $\mathcal{L}_k$ is the MAE on the validation dataset after the model is trained on $k$th inputs set.

In order to reduce variance of this estimation, we employ a baseline function $b$ \citep{zoph2016neural}:
\begin{eqnarray}
	\nabla_{\theta^v} J(\theta^v) \approx \frac{1}{n} \sum_{k=1}^{n}\sum_{i=1}^{T}\nabla_{\theta^v}\log P(c^v_i|\mathbf{x}^v_{i}; \theta^v) (e^{b-\mathcal{L}_k})
\end{eqnarray}
where $b$ is an exponential moving average of the previous MAEs on the validation set.

If we take the visual input gate controller as an example, the detailed training algorithm for the visual input gate controller is shown in Algorithm \ref{alg:rl}. The acoustic input gate is trained in the same manner.

\begin{algorithm}[H]
        \caption{\textsc{Train gate controller}} \label{alg:rl}
    \begin{algorithmic}[1]
        \Function{trainGateController}{$C^v$}
            \For{$epoch \gets 1 : epoch\_num$ }
                \For{$k \gets 1 : n$ }
        			\For{$i \gets 1 : T$ }
        				\State $p\_pass = predict(C^v, \mathbf{x}_i^v)$
                        \State $\mathbf{x}'^v_i \gets 0$
                        \State $\mathbf{x}'^v_i \gets \mathbf{x}^v_i$ with probability $p\_pass$ 
      				\EndFor
                    \State $loss_k \gets train{{\lasts }}(\mathbf{x}^w_{1:T}, \mathbf{x}^a_{1:T}, \mathbf{x}'^v_{1:T})$  
      			\EndFor
                \State $updateController(C^v, loss_k, loss\_baseline)$
                \State $updateLossBaseline(loss_k, loss\_baseline)$
      		\EndFor
        \EndFunction
    \end{algorithmic}
\end{algorithm}
\section{Experimental Methodology}
\label{Experimental Methodology}

In this section, we describe the experimental methodology including the dataset, data splits for training, validation and testing, the input features and their preprocessing, the experimental details and finally review the baseline models that we compare our results to.

\subsection{CMU-MOSI Dataset}
\begin{figure}[tb]
\begin{center}
\includegraphics[width=0.45\textwidth]{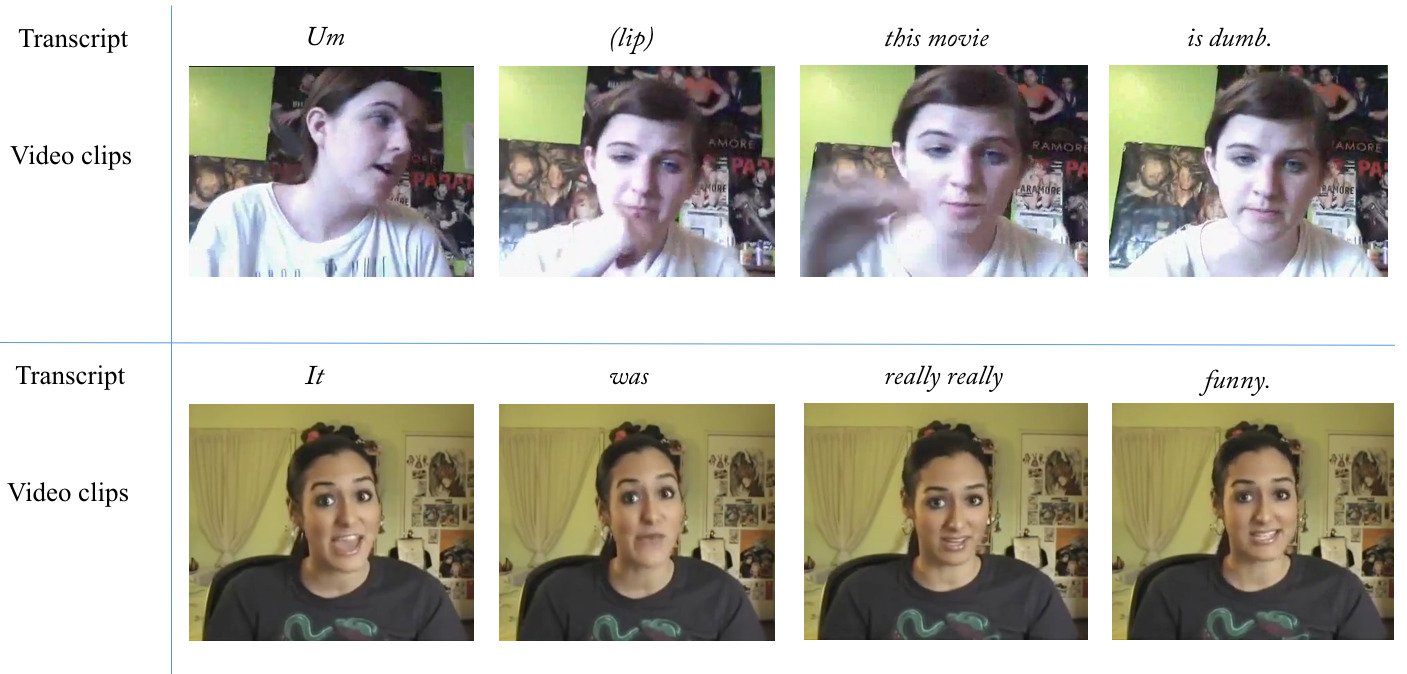}
\caption{A snapshot from the CMU-MOSI dataset, where text, visual and audio features are aligned. For example, in the bottom row of Figure \ref{mosi}, the first scene is labeled with text - the speaker is currently saying the word “It”, this is aligned with the video clip of her speaking that word where she looks excited.}
\label{mosi}
\end{center}
\end{figure}

We test on the Multimodal Corpus of Sentiment Intensity and Subjectivity Analysis (CMU-MOSI) dataset \citep{zadeh2016mosi}, which is a collection of online videos in which a speaker is expressing his or her opinions towards a movie. Each video is split into multiple clips, and each clip contains one opinion expressed by one or more sentences. A clip has one sentiment label $y \in [-3,+3]$ which is a continuous value representing speaker's sentiment towards a certain aspect of the movie. Figure \ref{mosi} depicts a snapshot from the CMU-MOSI dataset.

The CMU-MOSI dataset consists of 93 videos / 2199 labeled clips and training is performed on the labeled clips. Each video in the CMU-MOSI dataset is from a different speaker. We use the train and test sets defined in \citep{wang2016select} which trains on 52 videos/1284 clips (52 distinct speakers), validates on 10 videos/229 clips (10 distinct speakers) and tests on 31 videos/686 clips (31 distinct speakers). There is no speaker dependent contamination in our experiments so our model is generalizable and learns speaker-independent features.

\subsection{Input Features}
We use text, video, and audio as input modalities for our task. 
For text inputs, we use pre-trained word embeddings (glove.840B.300d) \citep{pennington2014glove} to convert the transcripts of videos in the CMU-MOSI dataset into word vectors. This is a 300 dimensional word embedding trained on 840 billion tokens from the common crawl dataset. For audio inputs, we use COVAREP \citep{degottex2014covarep} to extract acoustic features including 12 Mel-frequency cepstral coefficients (MFCCs), pitch tracking and voiced/unvoiced segmenting features, glottal source parameters, peak slope parameters and maxima dispersion quotients. For video inputs, we use Facet \citep{emotient} and OpenFace \citep{baltruvsaitis2016openface,zadeh2017convolutional} to extract a set of features including facial action units, facial landmarks, head pose, gaze tracking and HOG features \citep{zhu2006fast}. 

\subsection{Implementation Details}
Before training, we select the best 20 features from Facet and 5 from COVAREP using univariate linear regression tests. The selected Facet and COVAREP features are linearly normalized by the maximum absolute value in the training set. 

For the {\lasts } model, we set the number of hidden units of the LSTM as 64. The maximum sequence length of the LSTM, $T$, is 115. There are 50 units in the ReLU fully connected layer. The model is trained using ADAM \citep{DBLP:journals/corr/KingmaB14} with learning rate 0.0005 and MAE (mean absolute error) as the loss function. 

For the {\ours } model, the visual and audio controllers are each implemented as a neural network with one hidden layer of 32 units and sigmoid activation. The number of samples $n$ generated from the controller at each training step is 5. Each sampled {\lasts } model is trained using ADAM \citep{DBLP:journals/corr/KingmaB14} with learning rate 0.0005 and MAE (mean absolute error) as the loss function. The input gate controller is then trained using ADAM \citep{DBLP:journals/corr/KingmaB14} with learning rate 0.0001.

\section{Experimental Results}
\label{Results}

\subsection{Baseline Models}
\label{base}
We compare the performance of our methods to the following state-of-the-art multimodal sentiment analysis models:

\textbf{SAL-CNN} (Selective Additive Learning CNN) \citep{wang2016select} is a multimodal sentiment analysis model that attempts to prevent identity-dependent information from being learned so as to improve generalization based only on accurate indicators of sentiment. 

\textbf{SVM-MD} (Support Vector Machine Multimodal Dictionary) \citep{zadeh2016multimodal} is a SVM trained for classification or regression on multimodal concatenated features for each video clip. 

\textbf{C-MKL} (Convolutional Multiple Kernel Learning) \citep{DBLP:conf/emnlp/PoriaCG15} is a multimodal sentiment analysis model which uses a CNN for textual feature extraction and multiple kernel learning for prediction.

\textbf{RF} (Random Forest) is a baseline intended for comparison to a non neural network approach. 

\textbf{Random} is a baseline which always predicts a random sentiment intensity between $[3, -3]$ \citep{zadeh2016multimodal}. This is designed as a lower bound to compare model performance.

\textbf{Human} performance was recorded when humans are
asked to predict the sentiment score of each opinion segment \citep{zadeh2016multimodal}. This acts as a future target for machine learning methods.

Since sentiment analysis based on language has been well-studied, we also compare our methods with following text-based models:

\textbf{RNTN} (Recursive Neural Tensor Network) \citep{socher2013recursive} is a well-known sentiment analysis method that leverages the sentiment of words and their dependency structure.

\textbf{DAN} (Deep Average Network) \citep{iyyer2015deep} is a simple but efficient sentiment analysis model that uses information only from distributional representation of the words. 

\textbf{D-CNN} (DynamicCNN) \citep{kalchbrenner2014convolutional} is among the state of the art models in text-based sentiment analysis which uses a convolutional architecture adopted for the semantic modeling of sentences. 

Finally, any model with ``text'' appended denotes the model trained only on the textual modality of the CMU-MOSI video clips.

\subsection{Results}

In this section, we summarize the results on multimodal sentiment analysis. In Table \ref{table2}, we compare our proposed approaches with previous state-of-the-art multimodal as well as language-based baseline models for sentiment analysis (described in Section \ref{base}).

\begin{table}[tb]
\centering
\setlength\tabcolsep{4.0pt}
\caption{Sentiment prediction results on test set using different text-based and multimodal methods. Numbers are reported in binary classification accuracy (Acc), F-score and MAE, and the best scores are highlighted in bold (excluding human performance). $\Delta^{SOTA}$ shows improvement over the state-of-the-art. Results for RNTN are parenthesized because the model was trained on the Stanford Sentiment Treebank dataset \citep{socher2013recursive} which is much larger than CMU-MOSI.}
\begin{tabular}{l|l|l|l|l}
\hline
Modalities & Method         & Acc & F-score & MAE\\ \hline
\multirow{ 8}{*}{Text} & RNTN	\citep{socher2013recursive}					& (73.7) &   (73.4)      &   (0.990)  \\ \cline{2-5}
&DAN	 \citep{iyyer2015deep}					& 70.0 &   69.4     &   -   \\ \cline{2-5}
&D-CNN	\citep{kalchbrenner2014convolutional}					& 69.0 &   65.1     &     -   \\ \cline{2-5}
&SAL-CNN text    \citep{wang2016select}                   & 73.5 &   -      &   -      \\ \cline{2-5}
&SVM-MD text   \citep{zadeh2016multimodal}                   & 73.3 &    72.1     &   1.186      \\\cline{2-5}
&RF text    \citep{zadeh2016multimodal}                   & 57.6 &   57.5      &     -    \\ \cline{2-5}
&{\noatt } text (ours)						& 67.8 &    51.2   &   1.234  \\ \cline{2-5}
&{\lasts } text (ours)				& 71.3  &  67.3     &  1.062   \\ \hline
\multirow{ 10}{*}{Multimodal}&Random                      & 50.2 &   48.7      &     1.880    \\ \cline{2-5}
&SAL-CNN    \citep{wang2016select}                   & 73.0 &   -      &   -      \\ \cline{2-5}
&SVM-MD  \citep{zadeh2016multimodal}                     & 71.6 &    72.3     &   1.100      \\ \cline{2-5}
&C-MKL   \citep{DBLP:conf/emnlp/PoriaCG15}                   & 73.5 &    -    &   -     \\\cline{2-5}
&RF  \citep{zadeh2016multimodal}                      & 57.4 &   59.0      &     -    \\ \cline{2-5}
&{\noatt } (ours)         & 69.4 &   63.7    &   1.245      \\ \cline{2-5}
&{\lasts } (ours)         & 75.7 &   72.1     &   1.019      \\ \cline{2-5}
&{\ours } (ours)    & \textbf{76.5} & \textbf{73.4} & \textbf{0.955} \\ \cline{2-5}
&Human                       & 85.7 &   87.5      &    0.710     \\  \cline{2-5}
& $\Delta^{SOTA}$          & $\uparrow 3.0$ &   $\uparrow 1.1$     &    $\downarrow 0.145$     \\  \hline
\end{tabular}
\label{table2}
\end{table}

The multimodal section of Table \ref{table2} shows the performance of our two proposed approaches compared to other multimodal baseline methods. The model we proposed, {\ours } as well as the version without gated controller {\lasts }, both outperform multimodal and single modality sentiment analysis models. The {\ours } model gives the best result achieved across all models, improving upon the state of the art by 4.08\% in binary classification accuracy and 13.2\% in MAE. Since {\ours } is able to attend both in time, using soft attention as well as in input modality, using the \firstl Layer, it is not a surprise that this model outperforms all others.

The language section of Table \ref{table2} shows that {\lasts } on a single modality, language, obtains slightly worse performance than some language-based methods. This is because these methods use more complicated language models such as dependency-based parse tree. However, by combining cues from audio and video with careful multimodal fusion, {\ours } immediately outperforms all language-based and multimodal baseline models. This jump in performance shows that good temporal attention and multimodal fusion is key: our model benefit from the addition of input modalities more so than other models did.

\section{Discussion}
\label{Discussion}

In this section, we analyze the usefulness of our model's different components, demonstrating that the Temporal Attention Layer and the \firstl over input modalities are both crucial towards multimodal fusion and sentiment prediction.

\subsection{LSTM with Temporal Attention Analysis}

\begin{table}[tb]
\centering
\setlength\tabcolsep{6.0pt}
\caption{Sentiment prediction results on test set using {\noatt } model with different combinations of modalities. Numbers are reported in binary classification accuracy (Acc), F-score and MAE, and the best scores are highlighted in bold.}
\begin{tabular}{l|l|l|l|l}
\hline
Method & Modalities  & Acc & F-score & MAE\\ \hline
\multirow{ 6}{*}{\noatt} &text						& 67.8 &    51.2   &   1.234  \\ \cline{2-5}
& audio						& 44.9 &    61.9    &   1.511   \\ \cline{2-5}
& video					&  {}{44.9}  &    {}{61.9}     &    1.505    \\ \cline{2-5}
 & {}{text + audio}               & 66.8 &  55.3      &   \textbf{1.211}     \\ \cline{2-5}
& text + video               & 63.0 &   \textbf{65.6}     &   1.302    \\ \cline{2-5}
& text + audio + video            & \textbf{69.4} &  63.7     &   1.245     \\ \hline
\end{tabular}
\label{noatt}
\end{table}

\begin{table}[tb]
\centering
\setlength\tabcolsep{5.0pt}
\caption{Sentiment prediction results on test set using {\lasts } model with different combinations of modalities. Numbers are reported in binary classification accuracy (Acc), F-score and MAE, and the best scores are highlighted in bold.}
\begin{tabular}{l|l|l|l|l}
\hline
Method & Modalities  & Acc & F-score & MAE\\ \hline
\multirow{ 6}{*}{{\lasts }} & text				& 71.3  &  67.3     &  1.062   \\ \cline{2-5}
& audio						& 55.4 &   63.0     &   1.451   \\ \cline{2-5}
& video					& 52.3 &   57.3     &   1.443     \\ \cline{2-5}
& text + audio               & 73.5 &   70.3      &   1.036    \\ \cline{2-5}
& text + video               & {74.3} &    69.9    &   1.026     \\ \cline{2-5}
& text + audio + video           & \textbf{75.7} &   \textbf{72.1}     &   \textbf{1.019}      \\ \hline
\end{tabular}
\label{att}
\end{table}

\noindent \textbf{Language is most important in predicting sentiment.} In both the {\noatt } model (Table \ref{noatt}) and the {\lasts } model (Table \ref{att}), using only the text modality provides a better sentiment prediction than using unimodal audio and visual modalities.

\noindent \textbf{Acoustic and visual modality are noisy.} When we provide additional modalities to the LSTM model without attention (Table \ref{noatt}), the performance does not improve significantly. Using all three modalities actually leads to slightly worse performance in F-score and MAE as compared to using fewer input modalities. This allows us to deduce that the audio and video features are probably noisy and may hurt the model's performance if multimodal fusion is not carefully performed. 

\noindent \textbf{Temporal Attention improves sentiment prediction.} On the other hand, when we use the the {\lasts } model, Table \ref{att} shows that adding more modalities improves sentiment regression and classification. The {\lasts } (Table \ref{att}) consistently outperforms the {\noatt } (Table \ref{noatt}) across all modality combinations. We hypothesize that by using temporal attention, the model will assign the largest attention weights to time steps where all 3 modalities give strong, consistent sentiment predictions and abandon noisy frames altogether. As a result, temporal attention improves sentiment prediction despite the presence of noisy acoustic and visual modalities.

\noindent \textbf{Successful cases of the {\lasts } model.} To obtain a better insight into our model's performance, we provide some successful cases to demonstrate the contribution of the Temporal Attention Layer. By further studying the attention weights $\alpha$ in the {\lasts }, we can find which words/time steps the model focuses on. The following are examples of successful cases when we look at the textual modality alone. Each line represents a single transcript and the bold word indicates the word which the model assigns the highest attention weight to.

\begin{center}
\textit{I thought it was \textbf{fun}.\\
And she really \textbf{enjoyed} the film.\\ 
But a lot of the footage was kind of \textbf{unnecessary}.}
\end{center}
The highlighted words are all words that are good indicators of positive or negative sentiment.

\begin{figure}[tb]
\minipage{0.45\textwidth}
\begin{center}
\textit{He’s not gonna be looking like a chirper bright young man but early thirties really you \textbf{want} me to buy that.}
\end{center}
\endminipage \newline \newline
\minipage{0.5\textwidth}
\begin{center}
\includegraphics[width=0.8\textwidth]{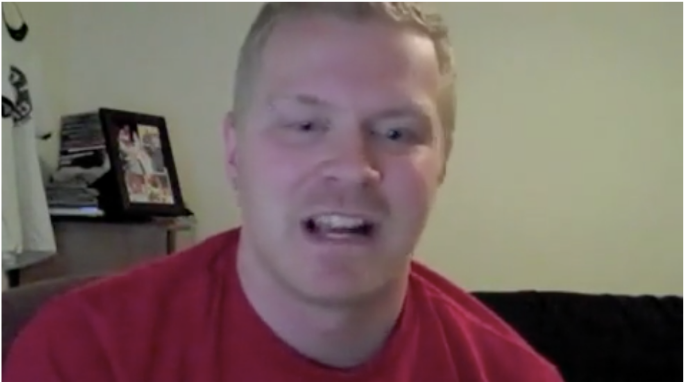}
\end{center}
\endminipage \newline \newline
\minipage{0.5\textwidth}
\begin{center}
Visual modality: Looks disappointed\\
{\noatt } sentiment prediction: \textbf{1.24}\\
{\lasts } sentiment prediction: \textbf{-0.94}\\
Ground truth sentiment: \textbf{-1.8}
\end{center}
\endminipage
\caption{Successful case 1: Although the highest weighted word extracted from the transcript (top) is ``\textit{\textbf{want}}'', with ambiguous sentiment, the {\lasts } leverages the visual modality (center), where the speaker looks disappointed, to make a prediction on video sentiment much closer to ground truth (bottom).}
\label{fig:g1}
\end{figure}

\begin{figure}[tb]
\minipage{0.45\textwidth}
\begin{center}
\textit{The only actor who can really sell their \textbf{lines} is Erin.}
\end{center}
\endminipage \newline \newline
\minipage{0.5\textwidth}
\begin{center}
\includegraphics[width=0.8\textwidth]{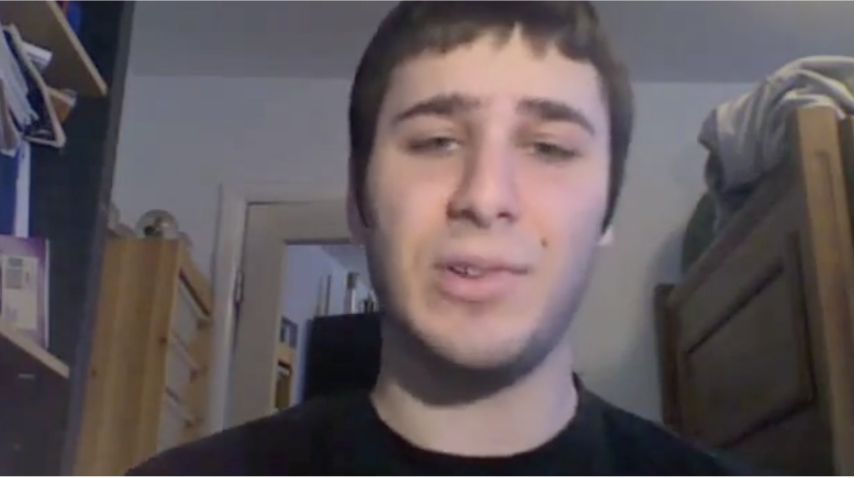}
\end{center}
\endminipage \newline \newline
\minipage{0.5\textwidth}
\begin{center}
Visual modality: Looks sad\\
{\noatt } sentiment prediction: \textbf{1.86}\\
{\lasts } sentiment prediction: \textbf{-0.3}\\
Ground truth sentiment: \textbf{-1.0}
\end{center}
\endminipage
\caption{Successful case 2: Although the highest weighted word extracted from the transcript (top) is ``\textit{\textbf{lines}}'', with ambiguous sentiment, the {\lasts } leverages the visual modality (center), where the speaker looks sad, to make a prediction on video sentiment much closer to ground truth (bottom).}
\label{fig:g2}
\end{figure}

\noindent \textbf{The {\lasts } model combines word meanings with audio and visual indicators.} Figure \ref{fig:g1} and Figure \ref{fig:g2} are examples where the {\lasts } model is successful when we use all 3 modalities. In these examples, the {\lasts } model is able to leverage the word level alignment of audio and visual modalities to overcome the ambiguity in the corresponding aligned word. The {\lasts } model is able to determine overall video sentiment to a greater accuracy as compared to the {\noatt } model without attention. 

\noindent \textbf{Word level fusion enables fine grained multimodal analysis.} We see that the model is indeed capturing the meaning of words and implicitly classifying them based on their sentiment: positive, negative or neutral. For neutral words, the model correctly looks at the aligned visual and audio modalities to make a prediction. Therefore, the model is learning the indicators of sentiment from facial gestures and tone of voice as well. This is a benefit of word level fusion since we can examine exactly what the model is learning at a finer resolution.

\subsection{\firstl Analysis}

\textbf{\firstl helps multimodal fusion.} The {\lasts } model is still susceptible to noisy modalities. Table \ref{all} shows that the {\ours} model outperforms the {\lasts } model on all metrics, indicating that there is value in attending in modalities using the \firstl. 

\begin{table}[tb]
\centering
\setlength\tabcolsep{4.5pt}
\caption{Sentiment prediction results on test set using {\noatt }, {\lasts } and {\ours } multimodal models. Numbers are reported in binary classification accuracy (Acc), F-score and MAE, and the best scores are highlighted in bold.}
\begin{tabular}{l|l|l|l|l}
\hline
Method & Modalities  & Acc & F-score & MAE\\ \hline
\multirow{ 1}{*}{\noatt} 
& text + audio + video            & 69.4 &  63.7     &   1.245     \\ \hline
\multirow{ 1}{*}{{\lasts }} 
& text + audio + video           & {75.7} &   {72.1}     &   {1.019}      \\ \hline
\multirow{ 1}{*}{{\ours }} 
& text + audio + video           & \textbf{76.5} &   \textbf{73.4}     &   \textbf{0.955}      \\ \hline
\end{tabular}
\label{all}
\end{table}

\begin{figure}[tb]
\minipage{0.45\textwidth}
\begin{center}
\includegraphics[width=0.9\textwidth]{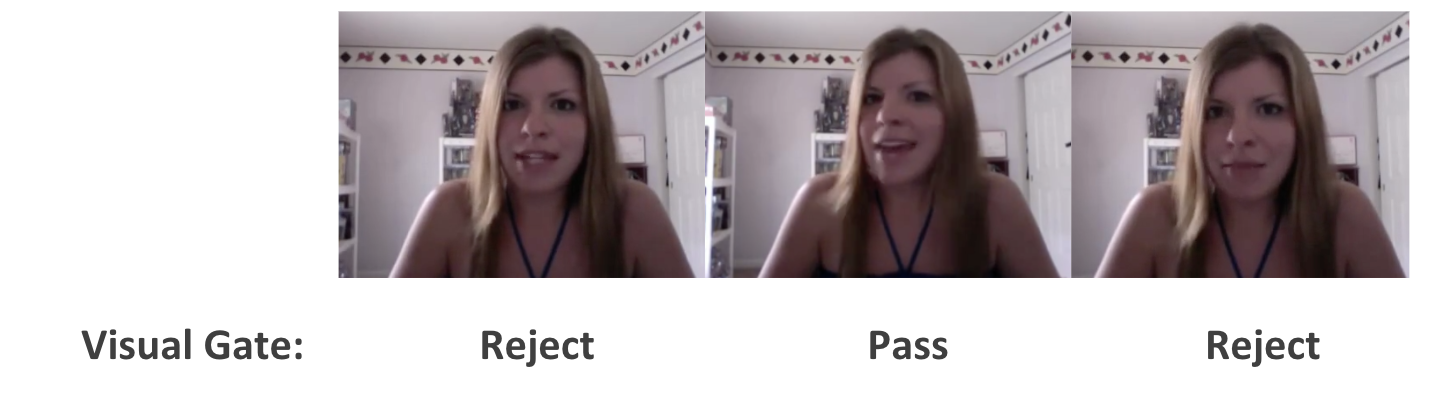}
\end{center}
\endminipage \newline \newline
\minipage{0.5\textwidth}
\begin{center}
{\lasts } sentiment prediction: \textbf{-2.00}\\
{\ours } sentiment prediction: \textbf{1.48}\\
Ground truth sentiment: \textbf{1.2}
\end{center}
\endminipage
\caption{Successful case 1: Across the entire video, the speaker’s facial features were rather monotonic except for one frame where she smiled brightly (left). Our visual input gate rejects the visual input at time steps before and after, but allows this frame to pass since the speaker is displaying obvious facial gestures. The prediction was much closer to ground truth as compared to without the input gate controller (right).}
\label{fig:good2}
\end{figure}

\noindent \textbf{{\ours } model correctly selects helpful modalities.} To obtain a better insight into the effect of the \firstl Layer, a successful example is shown in Figure \ref{fig:good2}, where the input gate controller for the visual modality correctly identifies frames where obvious facial expressions are displayed, and rejects those with a blank expression.

\begin{figure}[tb]
\minipage{0.48\textwidth}
\begin{center}
\textit{First of all I’d like to say \textbf{little} James or Jimmy he’s so cute he’s so ...}
\end{center}
\endminipage \newline \newline
\minipage{0.5\textwidth}
\begin{center}
\includegraphics[width=0.8\textwidth]{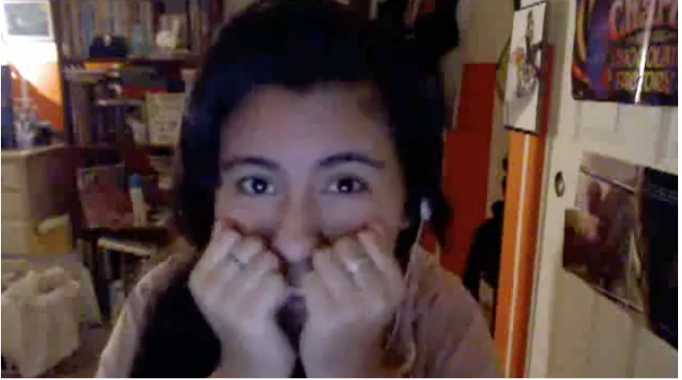}
\end{center}
\endminipage \newline \newline
\minipage{0.5\textwidth}
\begin{center}
Visual modality: Hands cover mouth\\
{\noatt } sentiment prediction: \textbf{1.23}\\
{\lasts } sentiment prediction: \textbf{-0.94}\\
{\ours } sentiment prediction: \textbf{1.57}\\
Ground truth sentiment: \textbf{3.0}
\end{center}
\endminipage
\caption{Successful case 2: The {\lasts } extracts the wrong word from the sentiment, extracting ``\textit{\textbf{little}}'' instead of the better word ``\textit{\textbf{cute}}'' (top). Upon inspection, the speaker is covering her mouth when the word ``\textit{\textbf{cute}}" is spoken (center), which leads to less attention weight on word ``\textit{\textbf{cute}}'' since the modalities are not consistently strong at that frame. As a result, the {\lasts } model makes a prediction on video sentiment that is further away from ground truth (bottom). However, the \firstl correctly rejects the noisy visual input at the time step of ``\textit{\textbf{cute}}'' (bottom). Including the \firstl improves the sentiment prediction back closer to ground truth.}
\label{fig:b2}
\end{figure}

\noindent \textbf{{\ours } model correctly rejects noisy modalities.} We now revisit a failure case of the {\lasts } model, where the speaker is covering her mouth during the word that gives best sentiment prediction, ``cute'' (Figure \ref{fig:b2}). The {\lasts } model is focusing on an uninformative time step and makes a poor sentiment prediction. In other words, the model may be confused if the added visual and audio modalities are uninformative or noisy. We found that the \firstl correctly rejects the noisy visual input at the time step of ``cute'' and the {\ours } model gives a sentiment prediction closer to the ground truth (Figure \ref{fig:b2}). This is a good example where the {\ours } model directly tackles the problem that motivated its development: the issue of noisy modalities that hurts performance when multimodal fusion is not carefully performed. Specifically, the {\ours } model was able to learn that the visual modality was mismatched with the textual modality, further recognizing that the visual modality was noisy while the corresponding word was a good indicator of positive speaker sentiment.

\section{Conclusion}
\label{Conclusion}

In this paper we proposed \ourl model for multimodal sentiment analysis. Our approach is the first of it's kind to perform multimodal fusion at word level. Furthermore to build a model that is suitable for the complex structure of speech, we introduce selective word-level fusion between modalities using gating mechanism trained using reinforcement learning. We use attention model to divert the focus of our model to important moments in speech. The stateful nature of our model allows for long interactions to be captured between different modalities. We show state of the art performance in MOSI dataset and we bring qualitative analysis of how our model is able to deal with various challenges of understanding communication dynamics.

\balance
\medskip
\small
\bibliographystyle{ACM-Reference-Format}
\bibliography{citations}

\end{document}